\documentclass[letterpaper, 10 pt, conference]{ieeeconf}  

\IEEEoverridecommandlockouts                              

\usepackage{multirow}
\usepackage{cite}
\usepackage{graphicx} 
\usepackage{subfigure}
\usepackage{hyperref}
\graphicspath{ {./fig/} }
\graphicspath{ {./fig/HD Map_viz/} }
\graphicspath{ {./fig/osm_viz/} }
\title{\LARGE \bf
OSM vs HD Maps: 

Map Representations for Trajectory Prediction 
}

\author{Jing-Yan Liao$^{\dagger}$, Parth Doshi$^{\dagger}$, Zihan Zhang$^{\dagger}$, David Paz$^{\dagger}$, Henrik Christensen$^{\dagger}$ 
\thanks{$^{\dagger}$Contextual Robotics Institute, University of California San Diego, La Jolla, CA 92093, USA
        }%
}

\begin{document}

\maketitle
\thispagestyle{empty}
\pagestyle{empty}

\begin{abstract}

While High Definition (HD) Maps have long been favored for their precise depictions of static road elements, their accessibility constraints and susceptibility to rapid environmental changes impede the widespread deployment of autonomous driving, especially in the motion forecasting task. In this context, we propose to leverage OpenStreetMap (OSM) as a promising alternative to HD Maps for long-term motion forecasting. The contributions of this work are threefold: firstly, we extend the application of OSM to long-horizon forecasting, doubling the forecasting horizon compared to previous studies. Secondly, through an expanded receptive field and the integration of intersection priors, our OSM-based approach exhibits competitive performance, narrowing the gap with HD Map-based models. Lastly, we conduct an exhaustive context-aware analysis, providing deeper insights in motion forecasting across diverse scenarios as well as conducting class-aware comparisons. This research not only advances long-term motion forecasting with coarse map representations but additionally offers a potential scalable solution within the domain of autonomous driving.

\end{abstract}

\section{INTRODUCTION}

Ensuring the safety of all road users is a fundamental mission of autonomous driving. To achieve this, self-driving systems must have a thorough understanding of their surroundings. In this context, motion forecasting emerges as a pivotal component, serving the purpose of modeling agent behavior within a scene and predicting their future trajectories. Motion forecasting models~\cite{hivt, lanegcn, varadarajan2021multipath} primarily rely on two key components: a map to depict the static elements in the environment, and past trajectories of agents within the scene, which account for the dynamic components. While past paths are typically well-defined as tracks, he pursuit of an optimal map representation~\cite{vectornet, bansal2018chauffeurnet} remains an ongoing challenge.

\begin{figure}[t]
        \centering
        \includegraphics[width=0.9\linewidth]{{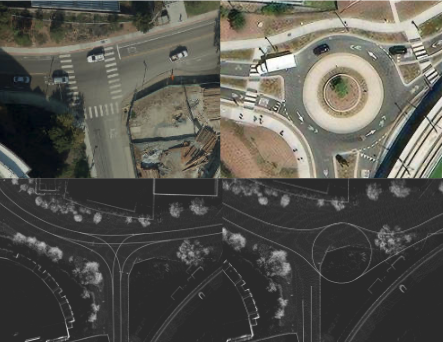}}
      \caption{Spatial Evolution of a road network at the UC San Diego Campus in both satellite image \cite{esri} and vector map perspectives \cite{avl}. The left side of the figure depicts the road network as of 2021, while and the  while the right side presents the current state of the road network. Below the satellite image, the corresponding vector map from the respective time period is displayed.}
      \label{roundabout_before_after}
\end{figure}

\begin{figure}[h]
        \centering
        \subfigure{\includegraphics[width=0.7\linewidth]{{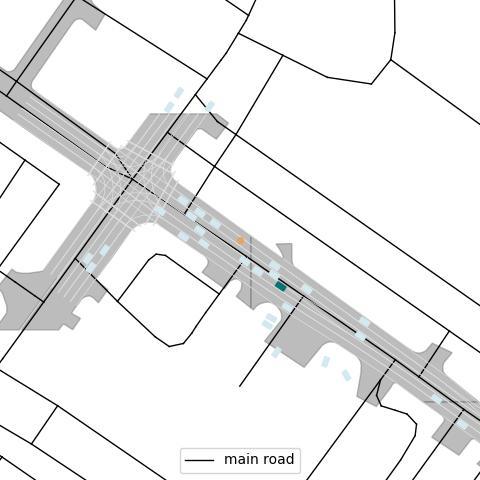}}}
      \caption{This figure illustrates the distinctions between HD Map and OSM representations. HD Map offers precise lane boundary delineations, lane connectivity (depicted by white lines), and driveable area (gray background). Conversely, OSM (indicated by black lines) primarily provides information on road connectivity, highlighting the contrast in map detail between the two representations.}
      \label{osm_ex}
   \end{figure}

HD Maps have emerged as the prevalent choice for map representation in motion forecasting applications due to their precision in conveying road markings, lane boundaries, and road geometry. This rich information contributes to the contextual and semantic understanding necessary for accurate predictions.  However, the accessibility to HD Maps is notably constrained in practice, not to mention  the considerable human and technological resources required for their creation and maintenance. This challenge hinders the large scale expansion of autonomous driving, especially in areas without up-to-date HD Map coverage. For instance, the UCSD on-campus autonomous driving practice consumes huge amount of human efforts on HD Map creation and maintaince. Unfortunately, this map quickly becomes outdated due to ongoing construction activities on campus, as depicted in Fig. \ref{roundabout_before_after}. Notably, not only do the locations of road lane markings change, but parts of the road network itself undergo significant alterations within a short timeframe, rendering the previously available HD Map obsolete and ineffective for autonomous driving applications. This example underscores the pressing need to explore alternative map representations that can accommodate dynamic real-world changes with increased efficiency and reduced manual labeling labor.
Interestingly, coarse map, which is widely used in localization \cite{osm_loc_1} \cite{osm_localization} and navigation \cite{osm_nav_1} \cite{osm_nav_2}for autonomous driving, has been somewhat overlooked as a potential map representation for motion forecasting due to reliance on HD Map until recently\cite{osm_first}. Open-source maps like OSM \cite{OpenStreetMap} and proprietary maps such as Google maps and Apple maps offer scalability advantages over HD Maps. Although they lack precise lane-level information, coarse maps include information about road connectivity and intersections as shown in Fig. \ref{osm_ex}, which hold substantial potential for motion forecasting. Although proprietary maps might contains more detailed as well as standardized representation of the roads, we choose to conduct experiments with OSM due to its open-source nature. 

Therefore in this work, we explore long horizon motion forecasting with OSM as a new map representation by utilizing the HiVT \cite{hivt} architecture as a basis. We specifically chose HiVT, HiVT-128 to be exact, for its state-of-the-art performance, and its real-time capability that scales well as the number of agents in the scene grows. Our key contributions are summarized as follows:

\begin{itemize}

\item \textbf{OSM for long tail forecasting: }We extend the application of OSM for the more challenging long-term motion forecasting. In contrast to previous research \cite{osm_av1}, which focused on short-term forecasting within the Argoverse1 Motion Forecasting Dataset \cite{argoverse1}, where the models were tasked with projecting 3 seconds into the future, our study evaluates the performance of our model using the Argoverse2 Motion Forecasting Dataset \cite{Argoverse2}. This dataset doubles the forecasting horizon, providing a more complex and realistic testing ground to assess the effectiveness of our OSM-based approach.
\item \textbf{Competitive performance of the OSM-based method: }Through the expansion of the receptive field and the incorporation of intersection flags, our OSM-based method achieves substantial improvements. Notably, we narrow the performance gap between HD Map-based models and our OSM-based model, demonstrating the competitive potential of our approach.
\item \textbf{Comprehensive Context-Aware Analysis: }We conduct an exhaustive context-aware comparison, both quantative and qualitative, considering both map representations and agent types. Our evaluation encompasses the visualization of motion forecasting outcomes across various scenarios, including straight lines, intersections, and curved roads. Furthermore, we delve into class-aware analysis, exploring how different map representations impact agent behavior under various map contexts.
\end{itemize}

By introducing OSM as a map representation and rigorously assessing HiVT's performance in diverse scenarios, this research provides a fresh perspective on long-term motion forecasting. It offers valuable insights into scalability in the realm of autonomous driving and beyond.

\section{Related Work}

\subsection{Map Representations for Motion Forecasting}

Initially, rasterized HD Maps \cite{bansal2018chauffeurnet,djuric2020uncertaintyaware,hong2019rules,salzmann2020trajectron,kamenev2022predictionnet} gained popularity because of the impressive results achieved by Convolutional Neural Networks (CNNs) in computer vision. Previous works  incorporate different features in the surrounding to provide rich and context-aware information for motion forecasting task. Some approaches entail the conversion of map elements (such as roads and crosswalks) into distinct layers with color-coded lane direction. Later, some studies have gone beyond simple rasterization to render more complex map features, including roadmaps, traffic lights, and speed limits, into bird's-eye view images. Afterward, rasterization techniques have been extended to encode semantic map details into a top-down spatial grid. This holistic approach aims to provide richer and more context-aware map representations, which are essential for enhancing the capabilities of motion forecasting models in making informed predictions. However, processing these rasterized representations with CNNs is computationally demanding and has limited receptive field, which could lead to higher error in longer term motion forecasting.

Recently, more research has shifted towards using vectorized HD Maps~\cite{shi2023mtr,gu2021densetnt,wayformer,gilles2022gohome,zhao2022trajgat} due to their more efficient representation. For instance, VectorNet \cite{vectornet} started this trend by sampling key points from lane splines to simplify the map and encode it with graph neural networks. LaneGCN \cite{lanegcn}, on the other hand, builds lane graphs with centerline segments and uses graph convolutional networks to capture information. HiVT transforms map elements into relative positions to guarantee translational invariance. TPCN \cite{tpcn} describes maps as ordered point sets and leverages a pointcloud based learning method to learn from the surrounding.

However, for coarse map, there is not a lot of research being done, especially for newer, longer horizon motion forecasting. Therefore, in this work, we reviewed more than 1000 scenarios in Argoverse 2 validation set, and cherry-picked some interesting examples and our finding from visualizing the prediction.

\section{Approach}

In this section, we first introduce OSM data format, and follow by presenting the methodology employed for extracting OpenStreetMap (OSM) data and its subsequent integration into the HiVT model. We delve into the specifics of OSM data formatting and elucidate the steps involved in the integration process.

\subsection{OSM Data format}
OpenStreetMap (OSM) serves as an extensive repository of diverse geographical attributes, encompassing various features such as road networks, structures, and facilities distributed across the Earth's surface. These geographical features are encapsulated within three fundamental data structures: nodes, ways, and relations.

Nodes, denoted as $n_i \in N$, encapsulate essential geographical information, including latitude, longitude, and a unique node identifier. Ways, represented as $w_i = \{ n_j\} \in W$, constitute aggregations of nodes that collectively define contiguous segments of roadways. Relations, articulated as $r_i = \{ n_j, w_k, r_l \} \in R$, serve to establish logical or geographic associations between disparate map objects. Each of these fundamental data structures accommodates supplementary metadata, such as road names or lane counts, stored as tags. This metadata enriches the core geographical information within the dataset.

For the purpose of our study, which focuses on motion forecasting, we selectively extract nodes and ways from the OSM data, omitting metadata. This decision is motivated by the inconsistency in label attributes across different geographic locations, an inherent characteristic of crowd-sourced data. However, it is worth noting that future research endeavors may explore methods to incorporate this metadata into our representation to fully harness the potential of OSM data.

\subsection{Incorporating OSM data into HiVT}
OSM, while sharing a graph-based structure with the Argoverse 2 HD Map, exhibits a sparser node distribution. Nevertheless, the inherently graph-based nature of OSM data makes it well-suited for integration with the HiVT model. The integration process can be summarized as follows:

\begin{enumerate}
  \item \textbf{Boundary Extraction:} Our process initiates with the acquisition of the boundary formed by all agent tracks within each Argoverse 2 scenario. The original agent coordinates, denoted as $a_i = (x_i, y_i)$, are initially provided in the city frame. We employ the Argoverse 2 API to transform these coordinates into the WGS84 coordinate system. This transformed boundary then serves as the basis for downloading relevant OSM map data.

  \item \textbf{Preprocessing for HiVT:} Following boundary extraction, we undertake several preprocessing steps to prepare the downloaded OSM data for integration into the HiVT model. Firstly, we transform all OSM nodes from the WGS84 coordinate system into the city frame, leveraging the Argoverse 2 API. Secondly, we perform interpolation on OSM ways to ensure uniformity in node-to-node distances. This interpolation employs a distance of 1.5 meters, aligning closely with the average centerline segment length observed in Argoverse 2. Finally, we transform the entire map into relative positions, similar to the HD Map preprocessing from HiVT. These preprocessing steps ensure compatibility and coherence between the OSM-based data and the HiVT model.
\end{enumerate}

This integrated approach facilitates the seamless incorporation of OSM data into the HiVT, enabling further analysis for long tail motion forecasting within urban landscapes.

\section{Experiments}
In this section, we present our comprehensive evaluation of the HiVT model on the publicly available Argoverse 2 Motion Forecasting Dataset.

\subsection{Dataset}

Initially, the HiVT model underwent evaluation using the Argoverse 1 dataset. However, for reasons outlined below, we opted to develop a tailored implementation for the Argoverse 2 dataset conversion. 

\begin{itemize}
    \item \textbf{Forecasting Horizon Expansion: } The Argoverse 2 dataset presents a significant departure from its predecessor, notably in terms of forecasting horizons. With each scenario extending to a duration of 11 seconds, the track history lengthens from 2 seconds to 5 seconds, while the forecasting horizon doubles to 6 seconds. This substantial increase in forecasting duration poses a more intricate challenge for motion forecasting, rendering it a valuable dataset for in-depth investigation.
    \item \textbf{Enhanced Class Information: } In contrast to Argoverse 1 with no class information provided, Argoverse 2 introduces a richer classification scheme comprising 10 non-overlapping classes, encompassing both static and dynamic agents. The inclusion of detailed class information pertaining to agents permits a more nuanced analysis of forecasting behavior, thereby enhancing our understanding of OSM-based model performance.
    \item \textbf{Diverse Scenarios: }The Argoverse 2 dataset is collected from six distinct cities, providing a diverse range of scenarios and environments. This geographical diversity allows us to assess the scalability and adaptability of the OSM-based method across varied urban landscapes.
\end{itemize}

By undertaking experiments on the Argoverse 2 Motion Forecasting Dataset, we aim to thoroughly examine the capabilities and performance of the HiVT model in addressing the challenges posed by extended forecasting horizons, enriched class information, and diverse real-world scenarios.

\subsection{Metrics}

We assess our model's performance using standard metrics for multi-modal motion forecasting: minimum Average Displacement Error (minADE), minimum Final Displacement Error (minFDE), and Miss Rate (MR). These metrics allow models to forecast up to 6 trajectories for each agent. The metric minADE quantifies the average l2 distance in meters between the best-predicted trajectory and the ground-truth trajectory across all future time steps, while minFDE measures the error specifically at the final future time step. MR represents the proportion of scenarios where the distance between the ground-truth endpoint and the best-predicted endpoint exceeds 2.0 meters.


\begin{table*}[h]
\caption{HiVT performance on Argoverse 2 validation set}
\label{table_av2}
\begin{center}
\resizebox{.8\linewidth}{!}{%
\begin{tabular}{|c c c c c c c|}
\hline
& receptive field  & minADE (m) & minFDE(m) & MR & Inference Speed (Hz) & VRam usage (MB)\\
\hline
\multirow{2}{5em}{HD Map} & 100 m & 0.943 & 1.934 & 0.287& 6.94 & 4404\\
 & 125 m & \textbf{0.929} & \textbf{1.876} &\textbf{ 0.277}& 7.05 & 4408\\
\hline
\multirow{2}{5em}{OSM} & 100 m & 1.375 & 3.234 & 0.433& 5.41 & 6266\\
 & 125 m & \textbf{1.043} & \textbf{2.241} & \textbf{0.324}  & 5.48& 6274\\
\hline
\multirow{1}{5em}{No Map} & N/A & 1.663 & 4.119 & 0.471 & 9.19 & 4384\\
\hline
\end{tabular}%
}
\end{center}
\end{table*}

\subsection{Implementation Details}
In our implementation, several key adjustments were made to enhance the compatibility of the HiVT model with the Argoverse 2 dataset and to leverage OpenStreetMap (OSM) data effectively.

\subsubsection{HiVT Model Enhancement}
We reconfigured some parts of the publicly available HiVT model to suit the requirements of the Argoverse 2 dataset. Specifically, we redesigned the dataloader tailored for Argoverse 2 data. Additionally, we modified the architecture to accommodate class information as an additional part of the input. This architectural adaptation allows the model to consider agent class information, enriching its contextual understanding and improving prediction accuracy.

\subsubsection{OSM Data Preprocessing}
On the OSM side, we performed crucial preprocessing steps to optimize the OSM data for integration with HiVT. This entailed interpolating the entire OSM graph to ensure that the distance between nodes uniformly averaged 1.5 meters. Furthermore, we employed a proximity-based approach to identify intersections, flagging nodes located within a 10-meter radius of specific markers such as stop signs and traffic lights. This attribute could provide contextual information for HiVT that improve intersection result. This process facilitated the inclusion of an \textit{is\_intersection} attribute, which plays a vital role in enabling the HiVT model to digest road networks information effectively.

\subsubsection{Training}
For both HD Map and OSM data, our model underwent training for 100 epochs on two RTX 3090 GPUs with a batch size of 32. We used the same optimizer, initial learning rate, weight decay, dropout rate, and learning rate decay as outlined in the original HiVT paper.

These implementation details are pivotal in achieving the desired synergy between the HiVT model and OSM data, ultimately enhancing the model's performance in the context of long-term motion forecasting on the Argoverse 2 dataset.

 \begin{figure}[htbp]
        \centering
        \subfigure{\label{fig:all_agent_DE}\includegraphics[width=0.8\linewidth]{{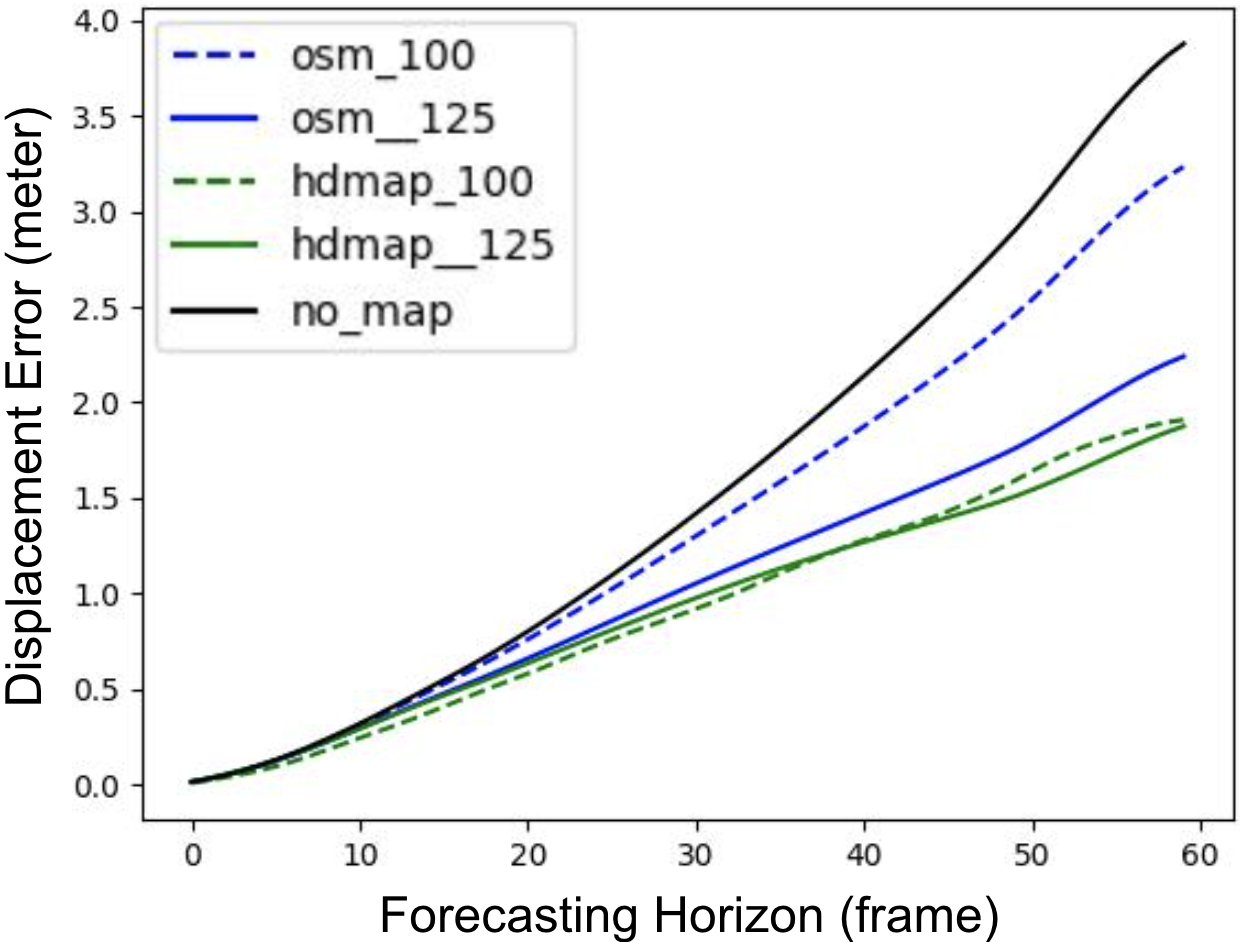}}}
      \caption{Comparing the OSM model with a receptive field of 100 meters (indicated by the blue dashed line) to one with a 125-meter receptive field (indicated by the blue solid line) reveals significant improvements in displacement error across the entire forecasting horizon. Additionally, OSM-125 exhibits remarkably similar performance to the HD Map-based method (green lines) within the first 30 frames into the future. The Argoverse 2 dataset provides data at a sampling rate of 10 Hz.}
      \label{all_agent_av2}
   \end{figure}
\subsection{Quantitative Analysis}
Our approach's validation results on the Argoverse 2 dataset are presented in Table \ref{table_av2}. Initially, when considering the original receptive field, which encompasses map information within a 100-meter radius around the agent, a notable performance gap becomes apparent between the HD Map-based method and our OSM-based approach. In this context, the OSM-based method only marginally outperforms scenarios where no map information is utilized.  It is worth noting that all the HiVT models used in this analysis are HiVT-128.

However, a pivotal transformation occurs when we expand the receptive field, which is the map information within certain radius around the agent, from 100 meters to 125 meters. Here, significant changes in performance dynamics emerge. While the HD Map-based results tend to plateau, the OSM-based method exhibits remarkable improvement. We attribute this divergence in performance to the limited learning power of our light-weight model HiVT, which may reach a performance bottleneck as additional map information provides diminishing returns for learning.  In contrast, OSM, having less detailed initial information, benefits significantly from an expanded receptive field, allowing HiVT to better capture the overall road network in the surrounding environment. Even with such a breakthrough in performance, we can still observe that there remains a gap in minFDE while the OSM-based method comes very close to matching the HD Map-based method in minADE. This highlights the need for more in-depth investigation.

To achieve a more comprehensive understanding, we conducted a frame-by-frame analysis of displacement errors, considering that tracked agent information is provided at a rate of 10 frames per second. This visual analysis is depicted in Figure \ref{all_agent_av2}. It is evident that with additional OSM information, the error remains constrained to an almost linear increase over time, even performing really close to HD Map-based methods, emphasizing the benefits of incorporating more map data for motion forecasting. Furthermore, increasing the receptive field is more feasible for the OSM-based method, as it is essentially impossible to exceed the range of available OSM data. Importantly, expanding the receptive field in HiVT incurs minimal computational overhead during inference. Our evaluation, conducted on a Titan Xp GPU with a batch size of 32, reveals negligible differences after the receptive field increase, both in terms of inference speed and GPU VRAM usage. This makes it an ideal choice for enhancing motion forecasting capabilities.

This quantitative analysis highlights the importance of the receptive field size at influencing the performance of our OSM-based approach compared to the HD Map-based method, revealing the potential for substantial enhancements in motion forecasting capabilities.

\begin{figure}
        \centering
        \subfigure{\includegraphics[width=\linewidth]{{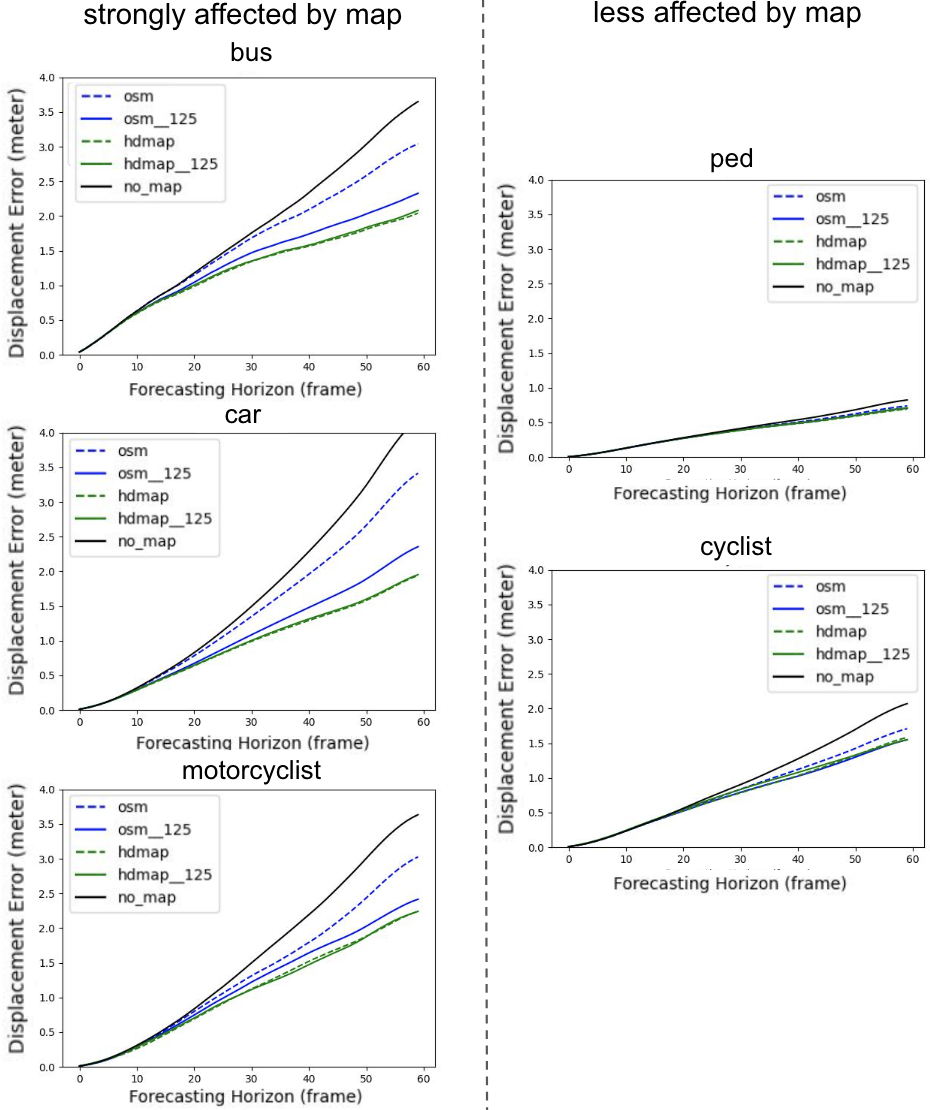}}}
      \caption{In this figure, two key conclusions are drawn. Firstly, for vehicles on the road (shown on the left), the use of HD Map provides a substantial advantage, suggesting that these classes benefit significantly from detailed lane-level information. Secondly, for pedestrians and cyclists (shown on the right), characterized by slow speeds and complex walking patterns, distinguishing between map representations in forecasting results is challenging, emphasizing the limited impact of map choice in such scenarios.}
      \label{all_class}
   \end{figure}

 \begin{figure}
        \centering
        \subfigure{\includegraphics[width=\linewidth]{{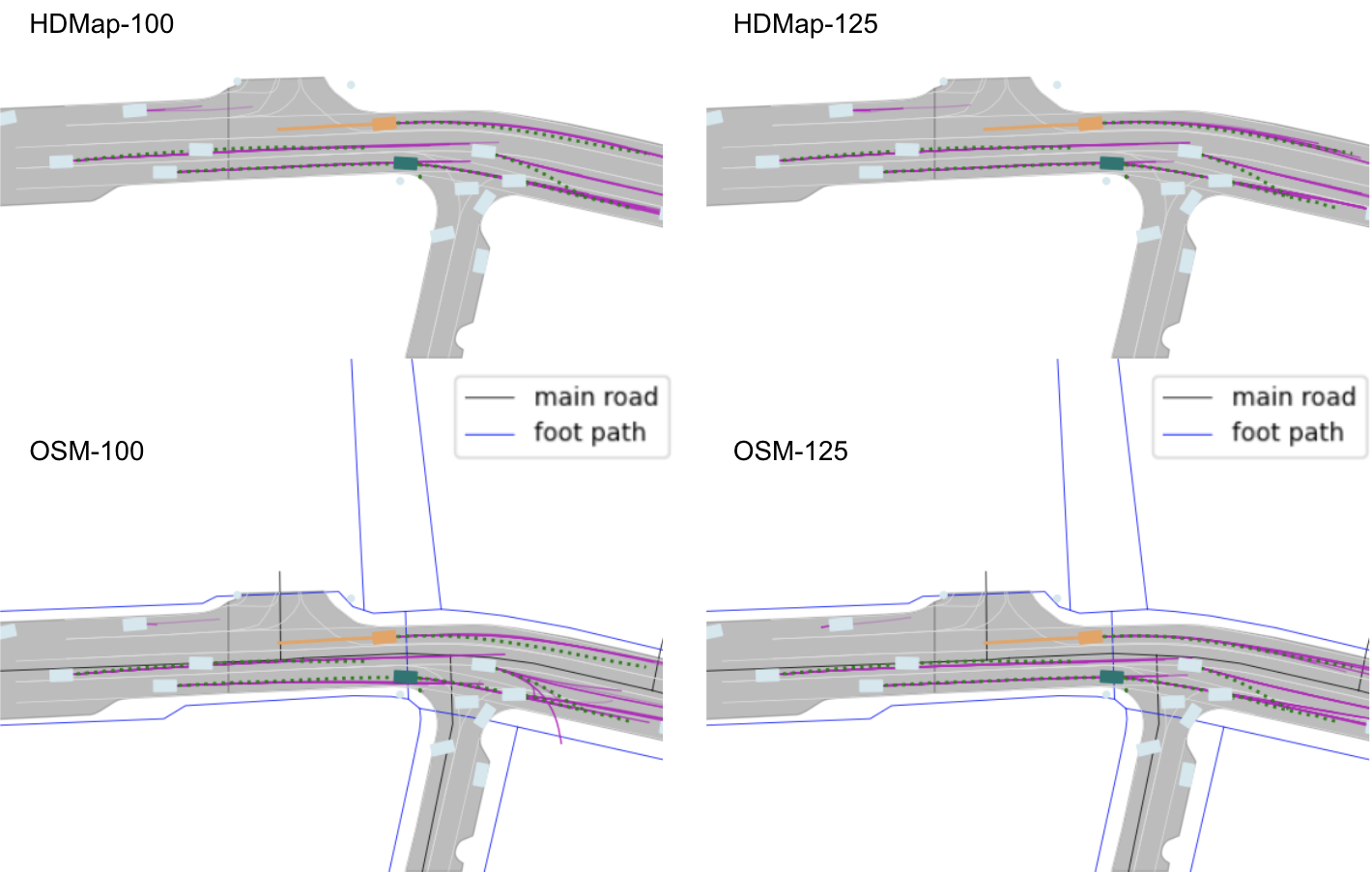}}}
      \caption{This figure illustrates a road segment featuring both straight and curved sections. The forecasting results are represented by the purple lines, where lower transparency indicates higher confidence. Remarkably, these forecasts closely align with the ground truth trajectories depicted by the green dotted lines. The orange box designates the focal agent, serving as the source for our metric evaluations.}
      \label{case_1}
   \end{figure}

 \begin{figure}
        \centering
        \subfigure{\includegraphics[width=\linewidth]{{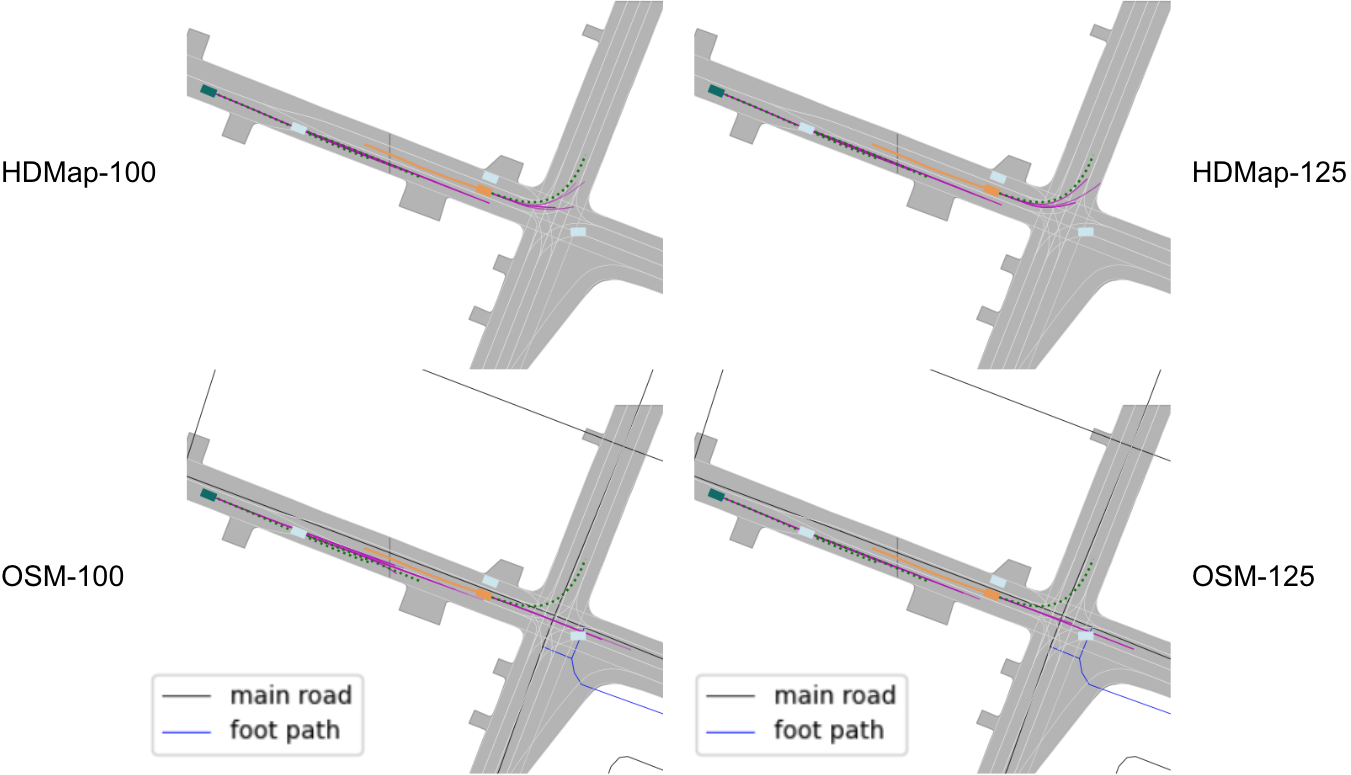}}}
      \caption{Illustration of an Intersection Scenario in HDMap and OSM-based Methods. In the HDMap scenario, the prediction of turning behavior is evident, which we attribute to the availability of lane-level information, as the vehicle is positioned to the left of the road. Conversely, in the OSM-based method, the absence of turning prediction is apparent due to the inconsistency in OSM labels, which resulted in the intersection not being labeled as such. This example highlights the significance of the \textit{is\_intersection} attribute for HiVT.}
      \label{case_3}
   \end{figure}

\subsection{Qualitative Analysis}
In addition to our quantitative findings, we conducted a qualitative analysis with a focus on class-specific performance and visualization, shown in Fig. \ref{all_class}, to gain deeper insights into the influence of map data on motion forecasting for different agent types.

In our class-specific analysis, we observed that the forecasting performance for pedestrians and cyclists showed limited improvement when  employing HD Map compared to OSM. This result aligns with expectations, as pedestrians often do not adhere to lanes while walking on sidewalks, resulting in both OSM and HD Map providing similar information in such cases. Conversely, classes such as cars, motorcyclists, and buses demonstrated noticeable performance enhancements when utilizing HD Map, emphasizing the potential significance of lane-level information in certain scenarios.

Nevertheless, we aim to uncover the underlying reasons for the performance gap between the OSM-based and HD Map-based methods. To achieve this, we categorized scenarios into three cases: straight roads, curved roads, and intersections. We then selected two representative scenarios from the Argoverse 2 validation set after a meticulous review of over 1000 scenarios. Drawing upon general knowledge, it was evident that lane-level information was primarily crucial at intersections for contextual understanding. To validate this hypothesis, we visualize scenarios such as straight and curved road in Fig.\ref{case_1} and intersection alone in Fig. \ref{case_3}. As expected, in both straight road and curved road scenarios, there was minimal difference in outcome between the OSM-based and HD Map-based methods. However, in the intersection scenario, where the inference of intersection locations was essential, it became evident that lane-level knowledge provided a significant contextual advantage. This observation hightlights the limitation of OSM-based method in motion forecasting given information we extracted from OSM.


\section{CONCLUSION}

In summary, this research presents an approach using OpenStreetMap (OSM) for long-term prediction when coupled with HiVT. Through the expansion of the receptive field and training our HiVT model with OSM data, we have achieved comparable results to HiVT trained with HD Map, as demonstrated on the Argoverse 2 dataset. Our exploration of map representations' impact on forecasting performance, with a focus on different agent types through our by-class analysis, has yielded valuable insights. Overall, our proposed methodology holds significant promise in advancing the scalability of autonomous driving systems by leveraging publicly accessible coarse map data. However, it is worth noting that the significance of lane-level information in motion forecasting remains crucial, especially in complex scenarios like intersections, where lane-level data continues to serve as a robust prior for predicting future trajectories.

\addtolength{\textheight}{-12cm}   







\bibliographystyle{unsrt}
\bibliography{ref}

\end{document}